\documentclass{Interspeech2024}

\interspeechcameraready

\title{Enhancing Speech-Driven 3D Facial Animation with Audio-Visual Guidance from Lip Reading Expert}

\name[affiliation={1\ast}]{Han}{EunGi}{}
\name[affiliation={2\ast}]{Oh}{Hyun-Bin}{}
\name[affiliation={2}]{Kim}{Sung-Bin}{}
\name[affiliation={3\dagger}]{Corentin}{Nivelet Etcheberry}{newline}
\name[affiliation={4}]{Suekyeong}{Nam}{}
\name[affiliation={4}]{Janghoon}{Joo}{}
\name[affiliation={1,2,5}]{Tae-Hyun}{Oh}{}

\address{
    $^1$Grad.~School of Artificial Intelligence and $^{2}$Dept.~of Electrical Engineering, POSTECH, Korea\\
    $^3$ENSC, Bordeaux INP, France \space{}
    $^4$KRAFTON, Korea \\
    $^5$Inst. for Convergence Research and Education in Advanced Technology, Yonsei University, Korea
}

\email{\{eungi, hyunbinoh, taehyun\}@postech.ac.kr}

\keywords{Speech-driven 3D Facial Animation, Audio-Visual Speech Recognition, Multimodal Perceptual Loss}

\usepackage{enumitem}
\usepackage{multirow}
\usepackage{wrapfig}
\usepackage{colortbl}
\usepackage[normalem]{ulem}
\usepackage{makecell}
\usepackage{subcaption}
\usepackage{hyperref}
\usepackage{url}

\usepackage[hang,flushmargin]{footmisc}

\setlist[itemize]{align=parleft,left=0pt,topsep=1mm,itemsep=0mm}

\definecolor{azure(colorwheel)}{rgb}{0.0, 0.5, 1.0}
\definecolor{nicegreen}{rgb}{0.0, 0.7, 0.1}
\definecolor{CuGray}{gray}{0.9}
\definecolor{rev}{rgb}{0.784, 0.003, 0.313}
\definecolor{pink}{cmyk}{0, 0.7808, 0.4429, 0.1412}
\definecolor{amethyst}{rgb}{0.6, 0.4, 0.8}
\definecolor{black}{rgb}{0.0, 0.0, 0.0}
\definecolor{tb3_yellow}{rgb}{0.996, 1.0, 0.6}
\definecolor{tb3_orange}{rgb}{0.980, 0.8, 0.604}
\definecolor{tb3_red}{rgb}{0.972, 0.6, 0.6}
\definecolor{blue}{rgb}{0.0, 0.0, 0.4}

\newcolumntype{g}{>{\columncolor{CuGray}}c}
\newcolumntype{z}{>{\columncolor{CuGray}}l}

\renewcommand{\paragraph}[1]{\noindent\textbf{#1.}\,\,}

\usepackage{xspace}

\def\onedot{.\@\xspace}
\def\eg{\emph{e.g}\onedot} 
\def\ie{\emph{i.e}\onedot}

\newcommand{\ba}{{\mathbf{a}}}

\newcommand{\bs}{{\mathbf{s}}}
\newcommand{\bt}{{\mathbf{t}}}

\newcommand{\bv}{{\mathbf{v}}}

\newcommand{\by}{{\mathbf{y}}}

\newcommand{\bA}{\mathbf{A}}

\newcommand{\bI}{\mathbf{I}}

\newcommand{\bL}{\mathbf{L}}

\newcommand{\bT}{\mathbf{T}}

\newcommand{\bV}{\mathbf{V}}

\newcommand{\bY}{\mathbf{Y}}

\newcommand{\calL}{{\mathcal{L}}}

\newcommand{\be}{\begin{eqnarray}}
\newcommand{\ee}{\end{eqnarray}}
\newcommand{\bee}{\begin{eqnarray*}}
\newcommand{\eee}{\end{eqnarray*}}

\newcommand{\matrixb}{\left[ \begin{array}}
\newcommand{\matrixe}{\end{array} \right]}

\newcommand{\para}[1]{\vspace{1mm}\paragraph{#1}}

\begin{document}

\maketitle

\begin{abstract}
    Speech-driven 3D facial animation has recently garnered attention due to its cost-effective usability in multimedia production.
    However, most current advances overlook the intelligibility of lip movements, limiting the realism of facial expressions.
    In this paper, we introduce a method for speech-driven 3D facial animation to generate accurate lip movements, proposing an audio-visual multimodal perceptual loss.
    This loss provides guidance to train the speech-driven 3D facial animators to generate plausible lip motions aligned with the spoken transcripts.
    Furthermore, to incorporate the proposed audio-visual perceptual loss, we devise an audio-visual lip reading expert leveraging its prior knowledge about correlations between speech and lip motions.
    We validate the effectiveness of our approach through broad experiments, showing noticeable improvements in lip synchronization and lip readability performance.
    Codes are available at \url{https://3d-talking-head-avguide.github.io/}.
\end{abstract}

\renewcommand{\thefootnote}{$^\ast$}
\footnotetext{These authors contributed equally.}
\renewcommand{\thefootnote}{$^\dagger$}
\footnotetext{This work is done during his student exchange program at POSTECH.}

\section{Introduction}
\begin{figure*}[t]
\centering
    \includegraphics[width=\linewidth]{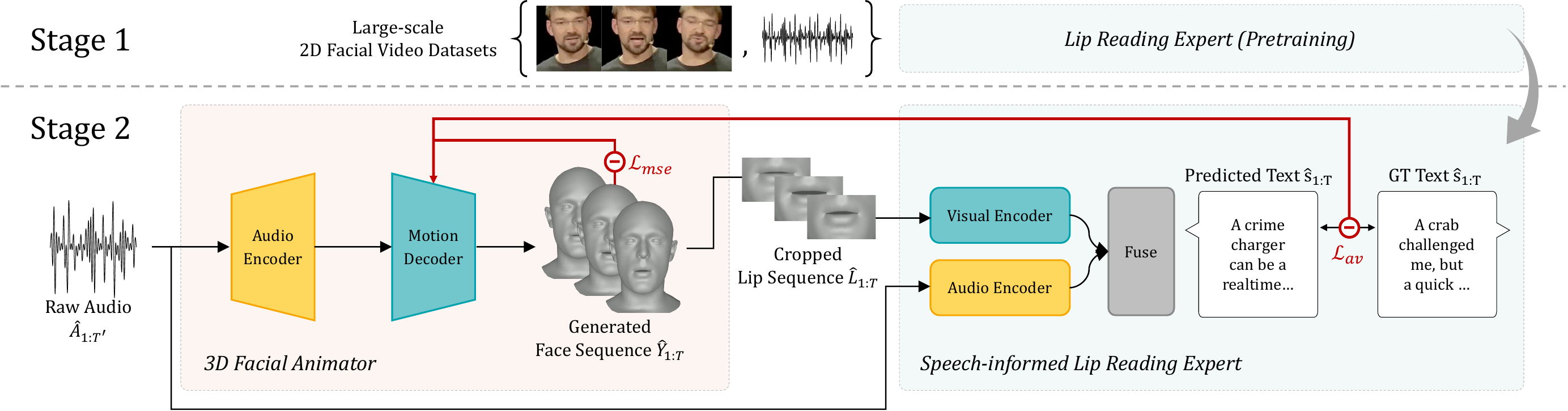}
    \vspace{-4mm}
    \caption{Overview of our proposed framework. We adopt the audio-visual lip reading expert~\cite{autoavsr} trained on the large-scale 2D datasets~\cite{lrs2,lrs3,avspeech,voxceleb2}
    and finetune it on 3D datasets~\cite{biwi,voca} concurrently with training the 3D facial animator.
    Given an input speech signal, a 3D facial animator regresses a sequence of 3D facial meshes and the following lip reading expert predicts the spoken transcript considering both the input speech signal and the sequence of lip regions of output faces.}
    \vspace{-3mm}
\label{fig:architecture}
\end{figure*}

The field of speech-driven 3D facial animation is growing rapidly, with a focus on generating realistic facial expressions from speech signals.
Animating 3D faces in practical applications often requires retouching or post-correction through manual intervention by skilled animators, which demands substantial human resources and costs.
In that sense, speech-driven 3D facial animation is widely receiving attention in industries such as entertainment, gaming, and virtual communication~\cite{edwards2020jali, taylor2017deep, wohlgenannt2020virtual, boyle2014narrative}, enhancing user experience and immersion.

There have been significant advancements towards adopting learning-based approaches in speech-driven 3D facial animation~\cite{voca,meshtalk,faceformer,codetalker,selftalk,emote,karras2017audio,laughtalk}; \eg, 3D facial movement generation is modeled by 1D convolutions over the temporal dimension (VOCA~\cite{voca}) or Transformer layers (FaceFormer~\cite{faceformer} and CodeTalker~\cite{codetalker}).
These methods focus on model architectures and exhibit appealing performance. 
Nonetheless, they primarily focus on minimizing Euclidean distance between ground truth and predicted mesh vertices, overlooking the importance of generating perceptually natural and intelligible lip movements, which is crucial for human visual comprehension.

To mitigate the unsatisfactory results, particularly in the lip region, we propose a method to guide the speech-driven 3D facial animation to better understand how the lip moves according to the spoken words, thus generating more plausible lip shapes.
Specifically, we introduce an audio-visual multimodal perceptual loss by leveraging a lip reading expert~\cite{autoavsr}, which incorporates both visual and speech inputs, to facilitate the speech-driven 3D facial animator to learn more speech-related information and generate plausible lip movements.
Furthermore, we implement the perceptual loss part in a two-stage training scheme. 
In the initial stage, we integrate the lip reading expert, which has been trained on extensive 2D talking face datasets~\cite{lrs2,lrs3,avspeech,voxceleb2}. 
Subsequently, in the second stage, we finetune the lip reading expert to optimize it on 3D facial datasets~\cite{biwi,voca}, concurrently with training a 3D facial animator.
This not only allows us to leverage prior knowledge of large 2D datasets but to reduce the domain gap from 3D face rendering.

While our proposed method is novel, there have been related attempts~\cite{wang2023seeing,spectre, selftalk} of using lip reading networks to enhance the intelligibility of lip shapes in their respective tasks. 
SelfTalk~\cite{selftalk} introduces a training framework that jointly trains a 3D face reconstruction module with a lip reading module, so that reconstructed 3D faces to be guided generate accurate transcripts synchronized with lip movements.
In contrast to our method, they rely on visual input to train the lip reading module, utilizing limited 3D datasets~\cite{biwi,voca}, without leveraging prior knowledge from large-scale 2D talking face datasets~\cite{lrs2,lrs3,avspeech,voxceleb2}.

SPECTRE~\cite{spectre}, which is closely related to our approach, demonstrates that leveraging a lip reading expert~\cite{ma2022visual} with prior knowledge can improve 3D facial reconstruction performance by minimizing the feature distance of lip movements between original and rendered videos. 
They also rely on a visual-only expert, while our method leverages speech and visual modalities together in our perceptual loss.

We conduct extensive experiments on BIWI~\cite{biwi} and VOCASET~\cite{voca} and show that our method is effective on different speech-driven 3D facial animator baselines (\ie, FaceFormer~\cite{faceformer} and CodeTalker~\cite{codetalker}) compared to the model without 2D prior knowledge or speech cues of the lip reading expert.
We further measure the quality of lip movements from lip readability's perspectives, such as Viseme Error Rate (VER) and Character Error Rate (CER) following SPECTRE~\cite{spectre}.

Our main contributions can be summarized as follows:
\begin{itemize}
    \item We propose an audio-visual perceptual loss, which guides the speech-driven 3D facial animator to learn more speech-related information and generate plausible lip movements.
    \item We devise an audio-visual lip reading expert tailored for the audio-visual perceptual loss, achieved via a two-stage training strategy: incorporating prior knowledge from extensive 2D talking face datasets in the initial stage, followed by fine-tuning of the lip reading expert on 3D talking face datasets.
\end{itemize}

\section{Method}
In this section, we describe the proposed method utilized in the 3D facial animator baselines (\ie, FaceFormer~\cite{faceformer} and CodeTalker~\cite{codetalker}) in detail.
The proposed framework consists of two components: a 3D facial animator and a speech-informed lip reading expert.
The 3D facial animator regresses a sequence of 3D face vertices from input speech signal, while the lip-reading expert maps the lip shape sequence, which is rendered with a differentiable face renderer, to textual representations.
The key idea of our method is to leverage the prior knowledge of the lip reading expert and to incorporate the audio modality to the expert for guiding the 3D face animation model to generate more plausible lip shapes.
Figure.~\ref{fig:architecture} illustrates the whole pipeline of our proposed framework.

\subsection{Speech-Driven 3D Facial Animator}
The 3D facial animator learns to regress a sequence of 3D facial movements from an input speech signal.
The regression process can be formulated as follows:
Let $\bY_{1:\bT} = (\by_1, ..., \by_\bT)$ denotes a sequence of ground truth 3D face vertices, where $\bT$ is the length of facial scan sequences and $\mathbf{\by}_t \in \mathbb{R}^{\bV \times 3}$ represents the face mesh of each frame which consists of $\bV$ vertices.
In addition, let $\bA_{1:\bT^{\prime}} = (\ba_1, ..., \ba_{\bT^{\prime}})$ be a sequence of speech representation, where $\bT^{\prime}$ is the length of the input speech signal.
The 3D facial animator predicts a sequence of 3D face vertices $\hat{\bY}_{1:\bT} = (\hat{\by}_1, ..., \hat{\by}_\bT)$ given a speech
signal $\bA_{1:\bT^{\prime}}$ as:
\begin{equation}
    \label{eq:talkinghead_step}
    \hat{\bY}_{1:\bT} = \text{FacialAnimator}_{\theta_1} (\bA_{1:\bT^{\prime}}),
\end{equation}
where $\theta_1$ denotes the weight of the 3D facial animator.
After generating the complete 3D facial motion sequence, the 3D facial animator is trained to predict accurate facial shapes by minimizing the Mean Squared Error (MSE) between the outputs of the 3D facial animator $\hat{\bY}_{1:\bT}$ and the ground truth $\bY_{1:\bT}$:
\vspace{-1mm}
\begin{equation}
    \label{eq:vert_loss}
    \calL_{\text{mse}} = \sum^{\bT}_{\bt=1} \sum^{\bV}_{\bv=1} \lVert \hat{\by}_{\bt, \bv} - \by_{\bt, \bv} \rVert^2.
\end{equation}

\subsection{Speech-Informed Lip Reading Expert}
The speech-driven 3D facial animators exhibit impressive performance in lip synchronization ability.
However, solely minimizing Euclidean distance between the ground truth and predicted face vertices is not sufficient to generate intelligible lip movements.
To generate realistic lip movements, we incorporate a powerful lip reading expert~\cite{autoavsr} that has the prior knowledge of the correlation between lip motions and their corresponding text content, which is learned from the extensive 2D talking face datasets~\cite{lrs2,lrs3,voxceleb2,avspeech}.
Specifically, we render the sequence of 3D face vertices from the 3D face animator into 2D video frames $\hat{\bI}_{1:\bT}$ with a differentiable face renderer as:
\begin{equation}
    \label{eq:rendering_step}
    \hat{\bI}_{1:\bT} = \text{Renderer} (\hat{\bY}_{1:\bT}).
\end{equation}
We crop the rendered gray-scale video frames $\hat{\bI}_{1:\bT}$ around the lip regions, resulting in the sequence of lip-cropped video frames $\hat{\bL}_{1:\bT}$. 
Then, the sequence of lip-cropped video frames $\hat{\bL}_{1:\bT}$ is fed to the lip reading expert.

Apart from SelfTalk~\cite{selftalk}, our speech-informed lip reading expert not only exploits prior knowledge from the 2D large-scale datasets but also incorporates both visual and speech information.
This approach produces facial mesh deformation that better corresponds to the speech, compared to models with a lip reading expert that only considers lip shapes (refer to Sec.~\ref{section:ablation}).
The lip reading expert predicts the transcript given both the sequence of lip-cropped video frames $\hat{\bL}_{1:\bT}$ and a sequence of speech representation $\bA_{1:\bT^{\prime}}$ as: 
\begin{equation}
    \label{eq:avsr_step}
    \hat{\bs}_{1:\bT} = \text{LipExpert}_{\theta_2}(\hat{\bL}_{1:\bT}, \bA_{1:\bT^{\prime}}),
\end{equation}
where $\theta_2$ is the weight of the lip reading expert.
Following \cite{autoavsr}, we incorporate the joint CTC/attention loss~\cite{ctc_attn_loss} into our objective function,
which penalizes the error between the predicted transcript $\hat{\bs}_{1:\bT}$ and ground truth $\bs_{1:\bT}$.
The CTC loss $\calL_{\text{ctc}}$ and the attention loss $\calL_{\text{ce}}$ are for learning the alignment between the predicted and actual transcripts, respectively.
Thus, the lip expert is finetuned with Audio-Visual (AV) perceptual loss $\calL_{\text{av}}$ which can be represented as a weighted sum of two losses:
\begin{equation}
    \label{eq:lipvertex}
    \calL_{\text{av}} = \lambda_{\text{ctc}}\calL_{\text{ctc}} + \lambda_{\text{ce}}\calL_{\text{ce}},
\end{equation}
where $\lambda_{\text{*}}$ denotes the loss weight, respectively.
We utilize the AV perceptual loss $\calL_{\text{av}}$ to guide the 3D facial animator to generate the output lip movements
that are comprehensible enough to guess the spoken words.

\subsection{Training Details}
\paragraph{The objective function}
As mentioned in SPECTRE~\cite{spectre}, na\"ively imposing the CTC loss to improve lip movement quality invokes face distortion since the model may prioritize achieving perfect lip reading recognition, a common phenomenon observed in adversarial attacks~\cite{akhtar2018threat, goodfellow2014generative}.
To address this issue, we add relative lip vertex loss $\calL_{\text{rlv}}$ as a regularizer which retains the spatial structure of the lip regions.
The relative lip vertex loss $\calL_{\text{rlv}}$ is calculated as the Mean Squared Error (MSE) between the output lip vertices and the ground truth lip vertices:
\vspace{-1mm}
\begin{equation}
    \label{eq:lip_vert_loss}
    \calL_{\text{rlv}} = \sum^{\bT}_{\bt=1} \sum^{\bV_L}_{\bv=1} \lVert \hat{\by}_{\bt, \bv} - \by_{\bt, \bv} \rVert^2,
\end{equation}
where $\bV_L$ denotes the number of lip region vertices.

To sum up, we train the facial animator and finetune the speech-informed lip reading expert with the objective function:
\begin{equation}
    \label{eq:loss_func}
    \calL = \lambda_{\text{mse}} \calL_{\text{mse}} + \lambda_{\text{av}} \calL_{\text{av}} + \lambda_{\text{rlv}} \calL_{\text{rlv}},
\end{equation}
where $\lambda_{\text{*}}$ denotes the loss weight, respectively.

\paragraph{Implementation details}
We conduct experiments on existing speech-driven 3D facial animation models, \ie, 
FaceFormer~\cite{faceformer} and CodeTalker~\cite{codetalker}, using them as a 3D facial animator in our framework.
Since CodeTalker separately trains the discrete motion prior and the auto-regressive facial animator, we train only the latter using pre-trained motion prior part.
We use the Adam optimizer and set the learning rate to 1e-4 without weight decay for 100 epochs.
Note that we reproduce all the 3D face animator baselines using publicly accessible codes and configurations.
We feed raw audio inputs into the Wav2Vec2.0~\cite{baevski2020wav2vec} encoder and extract the audio features from the last hidden state, following~\cite{faceformer,codetalker,selftalk}.
We adopt the same architecture of Auto-AVSR~\cite{autoavsr} as our lip-reading expert, which consists of a visual encoder, an audio encoder, a multi-layer perceptron (MLP), a projection layer, and a transformer decoder.
We train the lip reading expert with joint CTC/attention loss~\cite{ctc_attn_loss} on the LRS2~\cite{lrs2}, LRS3~\cite{lrs3}, AVSpeech~\cite{avspeech}, and Voxceleb2~\cite{voxceleb2} datasets, following the same training procedure of Auto-AVSR.
All experiments are conducted on a single NVIDIA A6000 GPU.

\section{Experiments}
\subsection{Experimental setup}

\subsubsection{Datasets}
We use two publicly available 3D datasets, VOCASET~\cite{voca} and BIWI~\cite{biwi}, for training and testing.
Both of them include the audio-3D scan pairs of utterances spoken in English.
We adopt the same data splits, \ie, training, validation, and test splits, following the FaceFormer~\cite{faceformer} and CodeTalker~\cite{codetalker}.

\para{VOCASET} 
VOCASET~\cite{voca} provides 480 audio-facial motion sequences for 12 subjects, captured at 60 Frames Per Second (FPS).
The dataset comprises 255 sentences, some of which are spoken by multiple speakers. The facial meshes are aligned with the FLAME~\cite{flame} topology, containing 5023 vertices.

\para{BIWI} 
BIWI~\cite{biwi} is a 3D audio-visual dataset with 40 unique sentences, which are all shared across 14 subjects and captured at a 25 FPS.
The dataset provides dynamic 3D face geometry aligned with 23,390 vertices. 
Each utterance is repeated twice with and without emotion, and we used the emotional subset.
There are 190 training sentences, 24 validation sentences, and two test datasets: BIWI-Test-A, containing 24 sentences from 6 subjects seen during training, and BIWI-Test-B, including 32 sentences from 8 unseen subjects.

\subsubsection{Evaluation metrics}
\paragraph{Lip synchronization} To measure lip synchronization
performance, we calculate Lip Vertex Error (LVE), which is a widely used metric for speech-driven 3D facial animation evaluation.
It computes the maximal L2 error by comparing all lip vertices of each predicted frame to the ground truth and takes the average over all frames in the test set.

\paragraph{Lip readability} LVE alone may not be enough to evaluate the lip movements, especially in the aspect of lip readability.
As a complement, we measured Character Error Rate (CER) and Viseme Error Rate (VER) between the actual and the predicted text from the lip reading expert.
VER is calculated by converting the predicted and actual text to visemes using the Amazon Polly phoneme-to-viseme mapping~\cite{amazonpolly}, following~\cite{spectre}.

\subsection{Quantitative Results}

\begin{table}[t]
    \centering
    \renewcommand*{\arraystretch}{0.85}
    \caption{Quantitative evaluation results on BIWI-Test-A.}
    \vspace{-2mm}
    \begin{tabular}{lccc} 
        \toprule
        Methods                      & \makecell{LVE $\downarrow$ \\ ($\times 10^{-4}$ mm)} & CER $\downarrow$  & VER $\downarrow$  \\
        \midrule
        FaceFormer~\cite{faceformer} & 6.0449           & 72.588\%          & 68.777\%          \\
        \textbf{+ AV Guidance}   & \textbf{5.5061}  & \textbf{68.423\%} & \textbf{62.422\%} \\ 
        \midrule
        CodeTalker~\cite{codetalker} &  5.3711          & 72.592\%           & 65.593\%           \\
        \textbf{+ AV Guidance}   & \textbf{4.8403}  & \textbf{70.711\%}  & \textbf{63.299\%}  \\
        \bottomrule
    \end{tabular}
    \vspace{-1.5mm}
    \label{tab:quantitative_biwi}
\end{table}

\begin{table}[t]
    \centering
    \renewcommand*{\arraystretch}{0.85}
    \caption{Quantitative evaluation results on VOCASET test split.}
    \vspace{-2mm}
    \begin{tabular}{lccc} 
        \toprule
        Methods                      & \makecell{LVE $\downarrow$ \\ ($\times 10^{-5}$ mm)} & CER $\downarrow$  & VER $\downarrow$  \\
        \midrule
        FaceFormer~\cite{faceformer} & 3.2496           & 76.244\%          & 66.932\%          \\
        \textbf{+ AV Guidance}   & \textbf{3.0987}  & \textbf{71.589\%} & \textbf{60.250\%} \\
        \midrule
        CodeTalker~\cite{codetalker} & 4.0557           & 75.988\%           & 65.105\%           \\
        \textbf{+ AV Guidance}   & \textbf{3.9884}  & \textbf{75.971\%}  & \textbf{64.912\%}  \\
        \bottomrule
    \end{tabular}
    \label{tab:quantitative_vocaset}
    \vspace{-3.5mm}
\end{table}

We incorporate our method into the 3D facial animator baselines (\ie, FaceFormer and CodeTalker) and calculate the Lip Vertex Error (LVE), Character Eror Rate (CER) and Viseme Error Rate (VER) for all sequences in the BIWI-TEST-A and VOCASET-Test datasets.
As shown in Table.~\ref{tab:quantitative_biwi} and \ref{tab:quantitative_vocaset},
Audio-Visual Guidance (AV Guidance), which includes 1) prior knowledge from the extensive 2D talking face dataset, 2) the relative lip vertex loss, and 3) speech information into the lip reading expert, improves all the evaluation metrics compared to the 3D facial animator baselines on both datasets.
This indicates that our method helps to generate intelligible lip movements.
In particular, the LVE on the BIWI-Test-A is 10$\%$ lower than the baselines, which shows the effectiveness of our proposed audio-visual perceptual loss.

\subsection{Qualitative Results}

\begin{figure}[t]
\centering
\includegraphics[width=\linewidth]{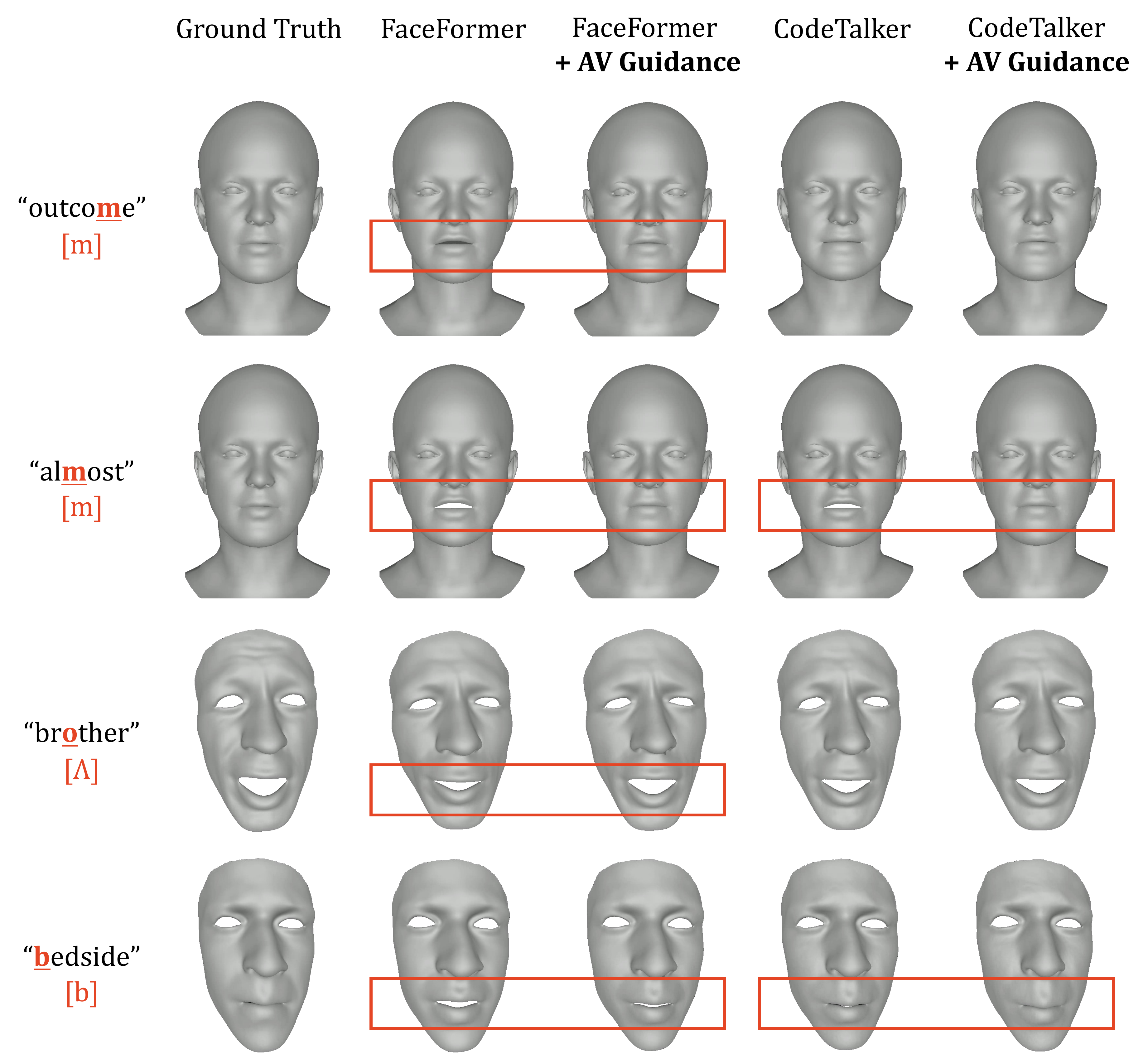}
\caption{Qualitative comparisons of output facial movements on VOCASET and BIWI. Compared to the 3D facial animator baselines~\cite{faceformer,codetalker}, the outputs of our method show better lip synchronization quality for both lip closure and opening words.}
\label{fig:qualitvative}
\end{figure}

We also conduct qualitative evaluation to assess the effectiveness of our method. 
Figure.~\ref{fig:qualitvative} illustrates visual comparisons between the ground truth, FaceFormer~\cite{faceformer}, CodeTalker~\cite{codetalker}, FaceFormer with AV Guidance, and CodeTalker with AV Guidance on both the VOCASET test split and BIWI-Test-A.
To ensure fair comparisons, we provide all models with the same speaking style as conditional input.
To evaluate the lip synchronization quality, 
we show several frames of output facial animations generated from the same audio input for each method.
We observe that our proposed methods produce more accurate lip closure movements compared to the baselines, accurately representing fully closed lip motions, particularly for syllables such as ``/m/'' or ``/b/''. 
Additionally, our method exhibits improved lip synchronization on mouth-opening as well, such as the syllable ``/\textipa{\textturnv}/''.
These comparison results underscore the effectiveness of employing an AV Guidance in achieving intelligible lip motions and accurately discerning various pronunciations.

\subsection{Ablation Study}
\label{section:ablation}
\begin{figure}[t]
    \centering
    \vspace{-2.5mm}
    \includegraphics[width=\linewidth]{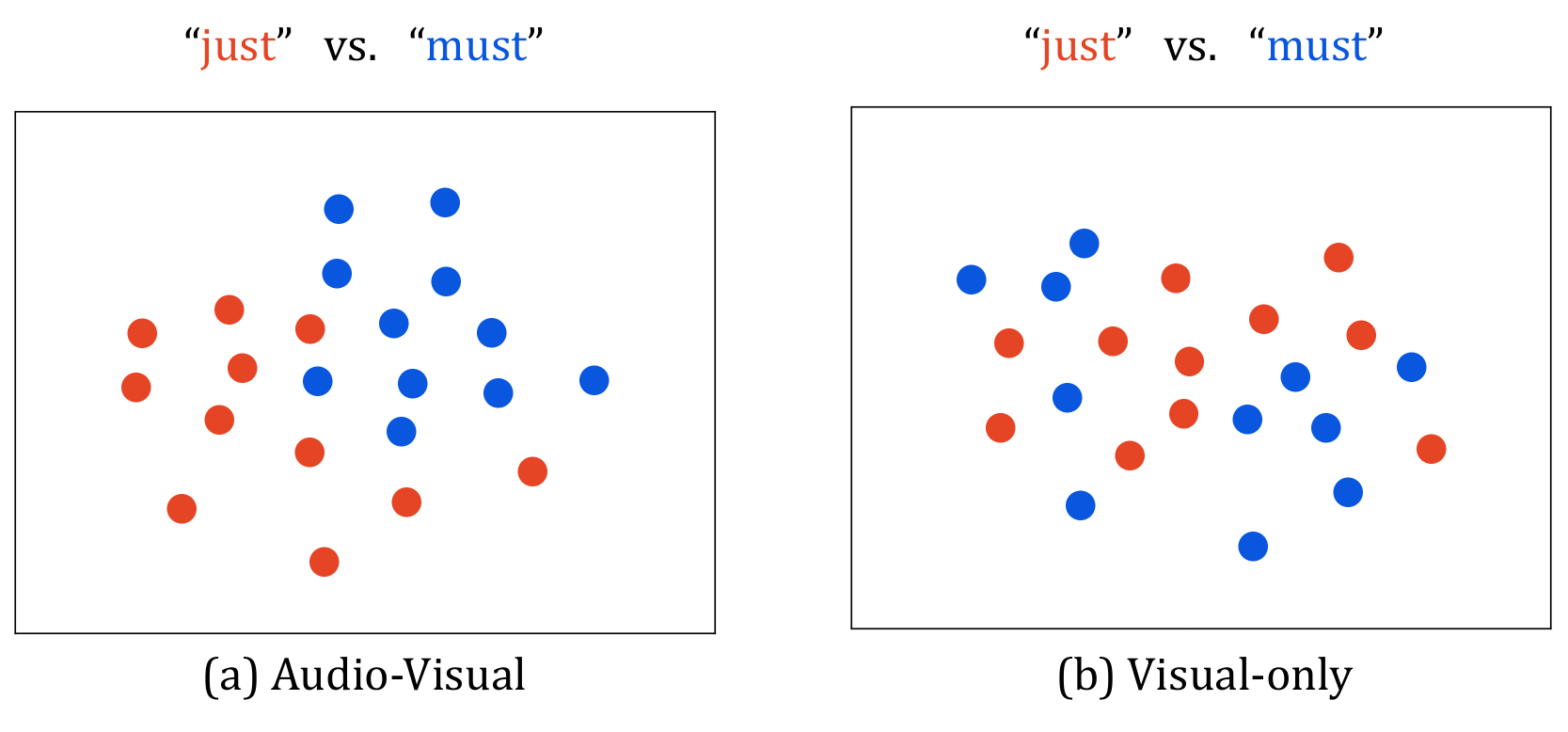}
    \vspace{-6mm}
    \caption{t-SNE visualization for features of audio-visual/visual-only lip reading expert.
    Distinct separation of features for words ``just'' (red) and ``must'' (blue) is observed in the audio-visual lip reading expert. However, with visual-only input, features become entangled, hindering distinction.}
    \label{fig:tsne_avsr_vsr}
    \vspace{-3mm}
\end{figure}

We conduct ablation studies to demonstrate the effectiveness of the different components of our method.
Specifically, we investigate the impact of AV Guidance, which comprises of prior knowledge of lip reading expert, the relative lip vertex loss, and the audio-visual perceptual loss.

\para{Impact of prior knowledge of lip reading expert} 
To investigate the effectiveness of the prior knowledge from the lip reading expert, we jointly train the 3D facial animator and the lip reading expert from scratch on the BIWI-Test-A dataset.

Table.~\ref{fig:ablation_faceformer} shows noticeable degradation in lip vertex error on baseline models~\cite{faceformer,codetalker} without prior knowledge.
These results indicate that the prior knowledge of lip reading expert plays an important role in generating more accurate lip motions, guided by the correlation between lip motion and spoken text in 2D talking face datasets~\cite{lrs2,lrs3,avspeech,voxceleb2}.

\para{Impact of relative lip vertex loss}
We investigate the impact of removing the relative lip vertex loss $\calL_{\text{rlv}}$ by optimizing the baseline with AV Guidance but excluding the $\calL_{\text{rlv}}$ loss.
Table.~\ref{fig:ablation_faceformer} show a deterioration in lip shape generation on both baseline models~\cite{faceformer, codetalker}, when removing the relative lip vertex loss.
This underscores the crucial role of this loss as a regularizer for retaining the spatial structure of the lip regions.
To assess the effectiveness of the relative lip vertex loss in AV Guidance regarding the improvement, we conduct an additional experiment on BIWI-Test-A. 
In this experiment, we optimize the baseline model with its original loss and the relative lip vertex loss.
FaceFormer with $\calL_{\text{rlv}}$ and CodeTalker with $\calL_{\text{rlv}}$ reveals $6.0976 \times 10^{-4}$mm and $5.2639 \times 10^{-4}$mm of LVE, respectively, showing subtle improvement or even decrease of performance compared to the baselines. 
This suggests that imposing more regularization
on the spatial structure of the lip regions, without the other components, is ineffective for accurate lip movements.

\begin{table}[t]
    \centering
    \renewcommand*{\arraystretch}{0.88}
    \caption{Ablation study for our components on BIWI-Test-A.}
    \vspace{-2mm}
    \begin{tabular}{lc} 
        \toprule
        Methods & LVE $\downarrow$ ($\times 10^{-4}$ mm) \\
        \midrule
        \makecell[l]{FaceFormer~\cite{faceformer} \textbf{+ AV Guidance}} & \textbf{5.5061} \\
        \quad w/o prior knowledge       & 5.9344 \\
        \quad w/o relative lip vertex loss $\mathcal{L}_{\text{rlv}}$ & 5.9023 \\
        \quad w/o speech information       & 6.0352 \\
        \midrule
        \makecell[l]{CodeTalker~\cite{codetalker} \textbf{+ AV Guidance}} & \textbf{4.8403} \\
        \quad w/o prior knowledge       & 5.4271 \\
        \quad w/o relative lip vertex loss $\mathcal{L}_{\text{rlv}}$ & 5.6524 \\
        \quad w/o speech information       & 5.3155 \\
        \bottomrule
    \end{tabular}
    \vspace{0mm}
    \label{fig:ablation_faceformer}
    \vspace{-3.5mm}
\end{table}

\para{Impact of audio-visual perceptual loss} 
We investigate how combining speech signals with lip reading enhances multimodal perceptual loss.
As shown in Table.~\ref{fig:ablation_faceformer}, leveraging speech information for predicting the spoken transcript leads to more effective learning signals related to speech information being transmitted to the 3D facial animator.
Since a single lip motion can correspond to multiple spoken texts, predicting the transcript from both visual and speech information yields better quality transcripts compared to predictions solely from visual information, \ie, lip motions. 
Consequently, with improved transcript prediction, the 3D facial animator is trained to generate more intelligible lip movements.
We visualize the features of the lip reading expert using t-SNE~\cite{tsne} in Figure.~\ref{fig:tsne_avsr_vsr}, illustrating cases of utilizing audio-visual information versus visual-only information.
Visual-only input entangles the features, hindering distinction of ``just'' (red) and ``must'' (blue). The audio-visual lip reading expert is shown to be able to guide lip movement.

\section{Conclusion}
In this paper, we introduce a method to guide the speech-driven 3D facial animation in comprehending lip movements corresponding to spoken words, thereby enhancing the realism of lip shapes. 
Our method proposes an audio-visual perceptual loss, which aids the speech-driven 3D facial animator in acquiring additional speech-related knowledge to produce plausible lip movements. 
We develop an audio-visual lip reading expert using a two-stage training approach: initial integration of prior knowledge from extensive 2D talking face datasets, followed by fine-tuning on 3D datasets. 
Extensive experiments show the effectiveness of our method in improving both lip synchronization and the intelligibility of generated lip motion, crucial aspects for human visual understanding.

\clearpage

\section{Acknowledgment}
This research was supported by a grant from KRAFTON AI.
This work was also partially supported by Institute of Information \& communications Technology Planning \& Evaluation (IITP) grant funded by the Korea government(MSIT) (No.RS-2023-00225630, Development of Artificial Intelligence for Text-based 3D Movie Generation; No.RS-2022-II220290, Visual Intelligence for Space-Time Understanding and Generation based on Multi-layered Visual Common Sense; No.RS-2019-II191906, Artificial Intelligence Graduate School Program (POSTECH))

\bibliographystyle{IEEEtran}
\bibliography{mybib}

\end{document}